%% file: acl_latex.tex
\pdfoutput=1

\documentclass[11pt]{article}

\usepackage[final]{acl}

\usepackage{times}
\usepackage{latexsym}

\usepackage[T1]{fontenc}

\usepackage[utf8]{inputenc}

\usepackage{microtype}

\usepackage{inconsolata}

\usepackage{graphicx}

\usepackage{amsmath}
\usepackage{amssymb}

\usepackage{makecell}
\usepackage{multirow}

\newcommand{\ModelName}{GigaCheck}
\title{\ModelName: Detecting LLM-generated Content via Object-Centric Span Localization}

\author{
  \textbf{Irina Tolstykh},
  \textbf{Aleksandra Tsybina},
  \textbf{Sergey Yakubson},
  \textbf{Aleksandr Gordeev},
\\
  \textbf{Vladimir Dokholyan},
  \textbf{Maksim Kuprashevich}
\\
\\
  SALUTEDEV LLC
\\
  \small{
    \textbf{Correspondence:} \href{mailto:irinakr4snova@gmail.com}{irinakr4snova@gmail.com}
  }
}

\begin{document}
\maketitle

\begin{abstract}
With the increasing quality and spread of LLM assistants, the amount of generated content is growing rapidly. In many cases and tasks, such texts are already indistinguishable from those written by humans, and the quality of generation continues to increase. At the same time, detection methods are advancing more slowly than generation models, making it challenging to prevent misuse of generative AI technologies. 
We propose \ModelName, a dual-strategy framework for AI-generated text detection. At the document level, we leverage the representation learning of fine-tuned LLMs to discern authorship with high data efficiency. At the span level, we introduce a novel structural adaptation that treats generated text segments as "objects." By integrating a DETR-like vision model with linguistic encoders, we achieve precise localization of AI intervals, effectively transferring the robustness of visual object detection to the textual domain.
Experimental results across three classification and three localization benchmarks confirm the robustness of our approach. The shared fine-tuned backbone delivers strong accuracy in both scenarios, highlighting the generalization power of the learned embeddings. Moreover, we successfully demonstrate that visual detection architectures like DETR are not limited to pixel space, effectively generalizing to the localization of generated text spans.
To ensure reproducibility and foster further research, we publicly release our source code.
\end{abstract}

\input{latex/1_Introduction}
\input{latex/2_RelatedWork.tex}

\input{latex/3_Method.tex}

\input{latex/4_Datasets.tex}
\input{latex/5_Experiments.tex}

\input{latex/6_Conclusions.tex}

\bibliography{custom}

\clearpage
\appendix
\input{latex/7_Appendix.tex}

\end{document}

%% file: latex/1_Introduction.tex
\section{Introduction}

\begin{figure*}[t!]
    \centering
    \includegraphics[width=\textwidth]{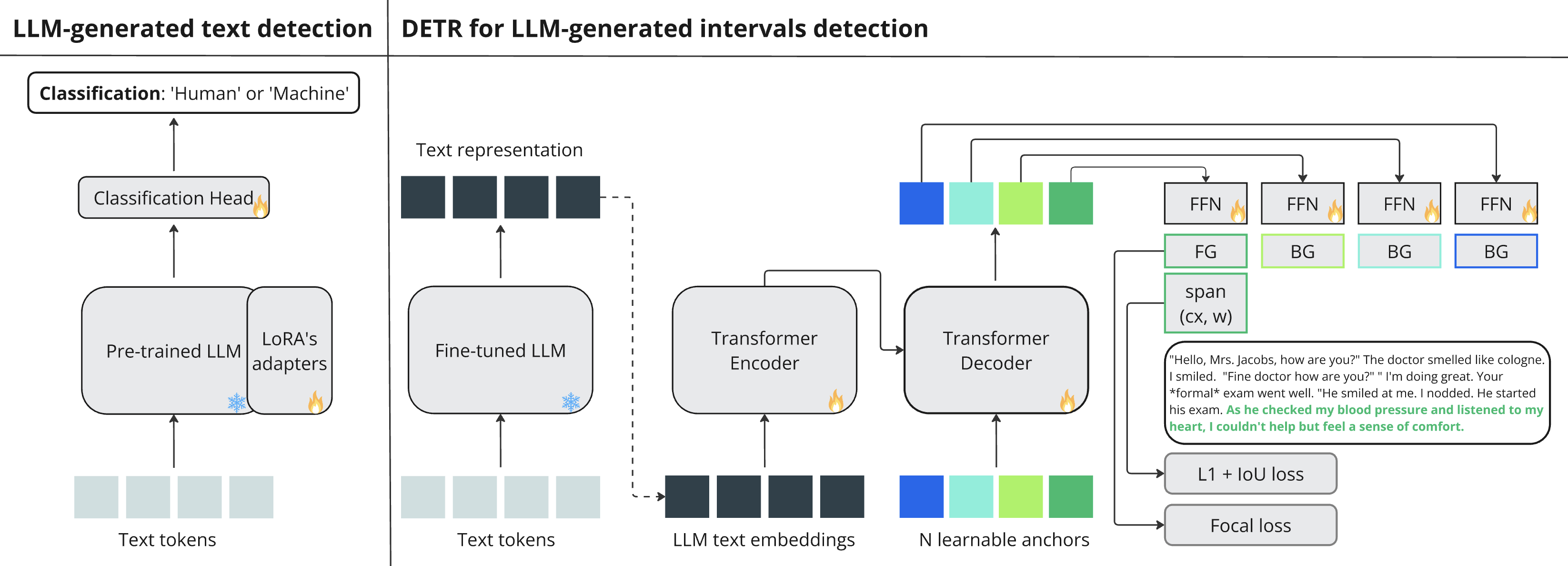}
    \caption{Overall architecture of \ModelName\ framework. Document-level detection is performed by fine-tuning an LLM. For span-level localization, we adopt a two-stage pipeline: (1) a fine-tuned LLM produces token embeddings, and (2) a detection transformer treats generated spans as objects and directly predicts character-level intervals. FG and BG denote the foreground and background labels assigned to each anchor.}
    \label{fig:Architecture}
\end{figure*}

The rapid development of Large Language Models (LLMs) has made their outputs difficult to distinguish from human-written text, raising concerns about the spread of spam and misinformation~\citep{mirsky2023threat,hanley2024machine}, fraud~\citep{grbic2023social,roy2023generating}, and academic cheating~\citep{stokel2022ai,kasneci2023chatgpt,perkins2023game,vasilatos2023howkgpt}. LLMs produce hallucinations~\citep{ji2023survey,thorp2023chatgpt} and outdated information, thereby spreading incorrect knowledge. Detecting LLM-generated content remains challenging, especially in mixed-authorship scenarios (Human-Machine collaborative texts), where existing document-level detectors lack sufficient reliability~\citep{liu2023detectability,wu2023survey}.

Recent approaches have shifted towards analyzing collaborative texts by identifying boundaries between sections of different authorship~\citep{zeng2024towards, zengdetecting, wang2023seqxgpt} or employing fine-grained token classification to extract spans~\citep{yin2025span}. 

In this paper, we propose a unified framework for generated text analysis, targeting both document-level classification and fine-grained span-level localization.
For the latter, we introduce a paradigm shift by reformulating text span detection as an \textit{object detection} problem. We employ a DETR-based architecture~\citep{DETR} that leverages representations from a fine-tuned LLM to predict character-based segments directly. Unlike previous sequence labeling methods that require manual post-processing to group tokens~\citep{kushnarevaai, zeng2024towards, wang2023seqxgpt}, our encoder-decoder transformer predicts continuous intervals end-to-end.

To keep the study focused and directly comparable with existing benchmarks, we limit this first investigation to English texts; adapting \ModelName\ to new languages is straightforward and left for promising future work.

To assess our approach, we adopt a two–step evaluation strategy. We begin with the challenging \emph{span-level localization} setting, demonstrating that the proposed DETR head can precisely pinpoint LLM-generated spans across three Human–Machine collaborative datasets. We then turn to three well-established binary-classification corpora. Although binary detection is less novel, these experiments verify that the very same LoRA-tuned backbone used by the DETR head learns embeddings that remain robust and discriminative for independent downstream tasks.

Our contributions are:

\begin{enumerate}

\item \textbf{Object Detection paradigm for text spans.}
To the best of our knowledge, DETR-style models have not yet been applied to locating intervals within natural language texts. We take this first step by adapting the architecture to detect LLM-generated segments as discrete objects, achieving strong results across three localization benchmarks. This approach eliminates the need for heuristic post-processing common in token-classification methods.

\item \textbf{Robust backbone for both detection and classification.}
The same LoRA-tuned backbone delivers state-of-the-art performance on three binary-classification datasets, proving that its embeddings transfer reliably between fine-grained localization and global document-level detection tasks.

\item \textbf{Open Source Availability.} 
To facilitate reproducibility and encourage future developments in the field, we make our source code publicly available at \url{https://github.com/ai-forever/gigacheck}.

\end{enumerate}

%% file: latex/2_RelatedWork.tex
\section{Related Works} \label{section:related_work}

\begin{table*}[h]
\centering
\setlength{\tabcolsep}{3pt}
\caption{Datasets used for training and evaluating the proposed approach 
(adapted from \citep{turingbench, tweepfake, zhang2024llm, mage, dugan2023real, kushnarevaai, zeng2024towards}). The tasks include both classification and detection. Note that “\#” represents “number of”.}
\small
\begin{tabular}{>{\centering\arraybackslash}m{1.0cm} p{2.0cm} p{2.8cm} p{5.5cm} >{\centering\arraybackslash}p{1.3cm} >{\centering\arraybackslash}p{2.0cm}}
\hline
\textbf{Task} & \textbf{Dataset} & \textbf{Generators} & \textbf{Domains} & \textbf{\# Texts} & \textbf{\# Boundaries} \\ 
\hline
\multirow{4}{*}{\rotatebox[origin=c]{90}{\textbf{Classification}}} 
& TuringBench & FAIR wmt20 & News & 17,163 & - \\
& TuringBench & GPT-3 & News & 17,018 & - \\
& TweepFake & Markov Chains, RNN, RNN+ Markov, LSTM, GPT-2 & Tweets & 25,572 & - \\
& MAGE & 27 LLMs from seven groups: GPT, LLaMA, GLM-130B, FLAN-T5, OPT, BigScience, EleutherAI & Reddit opinions, review, news, question answering, story, commonsense reasoning, Wikipedia paragraph, scientific writing & 447,674 & - \\
\hline
\noalign{\vskip 0.3ex}
\multirow{3}{*}{\rotatebox[origin=c]{90}{\textbf{Detection}}} 
& RoFT & GPT-2/XL, CTRL & Speeches, recipes, news, short stories & 8,943 & 1 \\
& RoFT-ChatGPT & GPT-3.5 Turbo & Speeches, recipes, news, short stories & 6,940 & 1 \\
& TriBERT & ChatGPT & Educational essays & 17,136 & 1-3 \\[1.5ex]
\hline
\end{tabular}
\label{tab:datasets}
\end{table*}

\subsection{Text Classification Methods}

Detecting machine-generated content has been widely studied. Work mainly focuses on binary classification (human vs. AI) \citep{zhang2024llm,liu2023detectability,bhattacharjee2024fighting,liu2023argugpt,uchendu2020authorship} and multiclass tasks to identify the specific generation model \citep{uchendu2020authorship,turingbench,uchendu2023toproberta,wang2024m4gt,detectgpt,wu2023llmdet}.

Statistical methods \citep{detectgpt, gltr, su2023detectllm, frohling2021feature} use metrics like entropy, perplexity, and n-gram frequency, and typically require access to the investigated LLMs. Neural-based approaches \citep{antoun2023towards,wang2024m4gt,guo2023close,liu2023check,zellers2019defending,solaiman2019release,uchendu2020authorship}, primarily using RoBERTa \citep{liu2019roberta}, provide more accurate results than statistical methods \citep{mage,liu2023check}, but lack robustness \citep{mage,koike2024outfox,krishna2024paraphrasing,chakraborty2023possibilities,tulchinskii2024intrinsic}. Recent works incorporate topological data analysis (TDA) \citep{uchendu2023toproberta,kushnareva2021artificial,tulchinskii2024intrinsic} or leverage LLMs as detectors. The authors of \citet{bhattacharjee2024fighting} apply GPT-3.5-turbo~\citep{chatgpt} and GPT-4~\citep{gpt4} models for the zero-shot binary classification task and demonstrate that both models have a very high misclassification rate. Our method extends neural-based detectors by fine-tuning an LLM to distinguish real and machine-generated text.

\subsection{Co-Written Text Analysis}
Several studies \citep{zhang2024llm,liu2023detectability} utilize neural-based classification models to classify Human-Machine collaborative texts. \citet{kushnarevaai} address the boundary detection task to determine where human-written text ends and machine-generated text begins, using fine-tuned RoBERTa and TDA-based time series. \citet{zeng2024towards} measure distances between adjacent segments to identify transitions, while \citet{zengdetecting} employ segmentation and classification of segments into AI-generated, human-written, or collaborative. A simpler approach by \citet{wang2023seqxgpt} identifies exact authorship for each sentence.

More recently, \citet{yin2025span} introduced Sci-SpanDet, a structure-aware framework designed specifically for scientific papers. They combine BIO-CRF sequence labeling with pointer networks to detect contiguous AI-generated spans, relying on section-specific contrastive learning that leverages the IMRaD structure (Introduction, Methods, Results, Discussion) of scientific documents. While effective in its target domain, Sci-SpanDet is inherently tied to structured document formats and cannot be directly applied to arbitrary texts lacking such explicit organization.

In contrast, our approach is domain- and structure-agnostic: by reformulating span detection as 1D object detection over character-level intervals, we eliminate the dependency on predefined document layouts or
sentence-level granularity, enabling flexible
detection of multiple generated segments in
any text.

\subsection{Transformer-based detection models}

DETR~\citep{DETR} is an end-to-end object detector based on transformers. DETR-like architectures have proven effectiveness in object detection \citep{zong2023detrs, hou2024relation, huang2022monodtr} and related tasks like video action detection  \citep{zhang2021temporal} and moment retrieval \citep{QVHighlights, moon2023correlation, 2410.01615}, where it is used to find temporal intervals in videos corresponding to a given text query. Inspired by these works, we propose to use a detection transformer model to perform span-level detection in texts.

Recent DETR modifications improve efficiency and accuracy: DeformableDETR~\citep{zhu2010deformable} speeds up convergence with deformable attention; DN-DETR~\citep{DN_DETR} uses denoising training to accelerate the training process and improve detection accuracy; DAB-DETR~\citep{liu2022dab} refines predictions by introducing learnable anchor boxes as DETR positional queries. DINO DETR~\citep{dino} combines these features and integrates RPN, while CO-DETR~\citep{zong2023detrs} enhances efficiency with auxiliary heads.

We adopt DN-DAB-DETR for its strong baseline and high localization accuracy~\citep{DN_DETR}. We also tested DAB-DETR, DeformableDETR, and CO-DETR, but DN-DAB-DETR consistently yielded the best results, so we adopt it throughout.

%% file: latex/3_Method.tex
\section{Methodology}
\label{section:method}

Figure~\ref{fig:Architecture} illustrates the architecture of \ModelName.
Our framework addresses two complementary tasks using a unified text-representation strategy: span-level localization and document-level classification. We employ a LoRA-tuned LLM whose token embeddings feed into two specialized heads. Below, we first present the backbone, followed by our novel object-centric span detector, and finally the classification head.

\subsection{Unified text-representation backbone}
\label{subsec:backbone}

We fine-tune a general-purpose decoder LLM, namely Mistral-7B,\footnote{\url{https://huggingface.co/mistralai/Mistral-7B-v0.3}} with LoRA~\citep{hu2021lora}.  
LoRA decomposes the weight matrix into two low-rank trainable matrices while keeping pre-trained weights frozen, yielding parameter-efficient fine-tuning (PEFT).  
We chose LoRA because (i) most of the datasets we use are small (see in Table~\ref{tab:datasets}), where PEFT often generalises better than full fine-tuning, and (ii) it converges much faster, saving GPU hours.  
Although results are reported with Mistral, the backbone is model-agnostic and any decoder-style LLM can be swapped in with minimal changes.

\paragraph{Proxy task.}
The LLM is tuned on a lightweight proxy classification task with two variants:

\begin{enumerate}
    \item \textbf{three-class proxy} (\emph{human}, \emph{machine}, \emph{collaborative}): used as a \emph{frozen feature extractor} for the DETR training.
    \item \textbf{two-class proxy} (\emph{human}, \emph{machine}): is \emph{trainable} along with the binary-classification head.
\end{enumerate}

\noindent For a document $\mathbf{X}$ we obtain tokens and embeddings via

\begin{equation}
\label{eq:eq1}
\begin{aligned}
& \mathbf{T}= \text{Tokenizer}(\mathbf{X}), \\   
& \mathbf{E}= \text{LLM}_{ft}(\mathbf{T}),\;\;
e_i\in\mathbb{R}^{d_{\text{model}}},
\end{aligned}
\end{equation}

\noindent where $\text{Tokenizer}$ is the BPE tokenizer shipped with Mistral and $\text{LLM}_{ft}$ is the LoRA-tuned required by the downstream head.
If fine-tuning is infeasible, pre-trained LLM embeddings may be substituted (see Appendix~\ref{apdx:pretrained_ft}).

\subsection{Object-centric Span Localization (DETR)}
\label{subsec:detr}

Our core contribution is the reformulation of text analysis as an object detection problem. We introduce a DETR-like head that treats LLM-generated segments as discrete objects, directly regressing 1-D character spans parameterized by $c$ and width $w$ (normalised to $[0,1]$). This approach avoids the limitations of token-level sequence labeling and operates independently of sentence boundaries.

\paragraph{Architecture.}
Embeddings $\mathbf{E}$, obtained in
Equation \ref{eq:eq1} from the frozen backbone, are first linearly projected to a lower dimension and then passed through a Transformer encoder to obtain contextual features:

\begin{equation}
\begin{aligned}
& \overline{\mathbf{E}}=\text{Linear}(\mathbf{E}),\quad \\
& \mathbf{R}= \text{TransformerEncoder}(\overline{\mathbf{E}})
 \end{aligned}
\end{equation}

\noindent We then follow DAB-DETR~\citep{liu2022dab}. 
A set of $N$ \emph{anchor-based learnable queries} $\mathbf{q}=\{q_0,\dots,q_{N-1}\}$ is initialised with reference points $(c,w)$, which act as initial hypotheses for the locations and lengths of LLM-generated spans. These queries are fed to the Transformer decoder, where sinusoidal encodings inject the anchor positions, and each cross-attention block concatenates positional and content embeddings, allowing the decoder to refine each anchor iteratively. 
At decoder layer $\ell$ the decoder predicts an offset $(\!\Delta c^{(\ell)},\Delta w^{(\ell)})$ for each anchor and updates it as
\[
(c,w)^{(\ell+1)}=(c,w)^{(\ell)}+(\!\Delta c^{(\ell)},\Delta w^{(\ell)}).
\]
After $L$ layers the decoder produces $N$ refined spans:

\begin{equation}
\label{eq:detr_decoder}
    \mathbf{o} = \text{TransformerDecoder}(\mathbf{q}, \mathbf{R}),
\end{equation}

\noindent where $\mathbf{o}=\{o_0,\dots,o_{N-1}\}$ corresponds one-to-one with the anchor queries.

\noindent As the model output, for each query the detector outputs a triplet $(c,w,p)$ comprising the refined centre c, width w, and a confidence score $p\!\in\![0,1]$ that the span is LLM-generated. Thresholding $p$ yields up to $N$ one-dimensional spans flagged as machine-written.

\noindent The number of queries $N$ is a dataset-level
hyperparameter set according to the maximum
expected span density.

\paragraph{Stabilising early training.}
As in DN-DETR \cite{DN_DETR}, the decoder is trained with two types of inputs:
(i) the learnable anchor queries, and (ii) noisy versions of the ground-truth (GT) spans. The model is trained to denoise these GT queries, while an attention mask prevents them from leaking information to the anchor queries.

\paragraph{Training loss.}
Before computing losses, we use Hungarian matching to pair each prediction with a GT span; the noised GT queries are excluded from this matching.  
The final objective is a weighted sum of L1, gIoU~\citep{generalized}, and Focal~\citep{lin2017focal} losses for the matched predictions, plus the same L1 and gIoU terms applied to the denoised GT queries.

\noindent We refer to the described detection transformer model as \textbf{\ModelName{}~(DN-DAB-DETR)}.

\subsection{Binary classification head}
\label{subsec:classifier}

The second head answers the document-level question  
\emph{“Is this text human-written or LLM-generated?”}.  
Formally, for a document $X$ we learn
\[
f_\theta: X \;\longrightarrow\; \{0,1\},\quad
f_\theta(X)=
\begin{cases}
0,&\text{human},\\
1,&\text{machine}.
\end{cases}
\]

\noindent We attach a two-layer MLP to the hidden state of the final \texttt{<EOS>} token of the \emph{two-class} LoRA variant and train it with binary cross-entropy.
The resulting model is referred to as \textbf{\ModelName{}~(Mistral-7B)}.

%% file: latex/4_Datasets.tex
\section{Datasets and Metrics} \label{section:bechmarks}

Table~\ref{tab:datasets} lists all datasets used in this work. 
We use the original train–test splits in Section \ref{section:experiments}, enabling comparison with other approaches trained on the same data.

\noindent\textbf{Classification datasets.} We evaluate the proposed approach for machine-written text classification using three datasets: TuringBench~\citep{turingbench}, TweepFake~\citep{tweepfake}, and MAGE~\citep{mage}. We prioritized these benchmarks while noting that other existing corpora, such as MixSet~\citep{zhang2024llm} or Ghostbusters~\citep{verma2023ghostbuster}, consist of a limited amount of data. Such small-scale datasets are known to be easily solvable and often fail to reflect the complexity of real-world detection scenarios~\citep{gritsai2024ai}. Regarding TuringBench, we specifically use the two subsets generated by FAIR~wmt20~\citep{chen2020facebook} and GPT-3~\citep{gpt3}, as these models produce texts most indistinguishable from human-written ones according to the dataset authors.

\noindent\textbf{Detection datasets.} We considered three datasets for Human-Machine collaborative text analysis, which have been created to address the task of identifying a boundary between human-written and machine-generated text: 
RoFT~\citep{dugan2023real}, RoFT-ChatGPT~\citep{kushnarevaai}, and TriBERT~\citep{zeng2024towards}. 

\noindent\textbf{Classification metrics.} We evaluate \ModelName\ as an LLM-generated content detector using classification accuracy (Acc), F1 score, AUROC, and average recall (AvgRec)~\citep{mage}, calculated as the average of recall scores for human-written (HumanRec) and machine-generated (MachineRec) texts.

\noindent\textbf{Detection metrics.} We use metrics such as sentence-wise MSE, Accuracy, and Soft Accuracy from \citet{kushnarevaai}, as well as a specialized form of the F1 score from~\citet{zeng2024towards}, to assess the quality of the model's predictions of the boundaries between sentences written by a human or an LLM. The authors of~\citet{zeng2024towards} consider 
$L_{topK}$, which represents the top-K boundaries identified by the algorithm, and $L_{Gt}$, which refers to the number of ground-truth boundaries. The F1 score is then determined using the following formula:

\begin{equation}
\label{eq:eq3}
F1@K=2\cdot\frac{|L_{topK}\cap L_{Gt}|}{|L_{topK}|+|L_{Gt}|}
\end{equation}
Further details on the calculation of each metric are provided in Appendix ~\ref{apdx:metrics}.

%% file: latex/5_Experiments.tex
\section{Experimental Results} \label{section:experiments}

In this section we first report span-detection results on three Human-Machine collaborative datasets, then present an extensive evaluation on three binary-classification benchmarks.
While the classification task itself is well studied, these additional experiments serve to verify that the proposed text-representation backbone produces embeddings that remain robust and discriminative for a separate downstream task.
Training details for all runs are provided in Appendix ~\ref{apdx:hyperparameters}.

\subsection{Detection Results}
\label{sec:detection}

To provide a comprehensive assessment, we benchmark \ModelName\ against a diverse spectrum of baselines operating at varying granularities.
We evaluate our span-detection method on the RoFT and RoFT-GPT datasets against approaches operating at the token level, sentence level, and document level. This inclusion allows us to compare our object-centric approach directly with traditional fine-grained methods.
For the TriBERT dataset, following established protocols, we compare our method with sentence-level approach.

\begin{table*}
  \centering\setlength{\tabcolsep}{4pt}
  \caption{Boundary detection results on RoFT and RoFT-ChatGPT datasets. The results for all methods, except ours, were taken from \citet{kushnarevaai}.}
  \small
  \begin{tabular}{lllllll}
  \hline
    \makecell{\textbf{Method}} & \multicolumn{3}{c}{\textbf{RoFT}} & \multicolumn{3}{c}{\textbf{RoFT-ChatGPT}} \\
     & Acc & SoftAcc1 & MSE & Acc & SoftAcc1 & MSE \\
    \hline
    RoBERTa + SEP~\citep{cutler2021automatic} & 0.50 & 0.80 & 2.63 & 0.55 & 0.79 & 3.06 \\
    RoBERTa~\citep{liu2019roberta} &  0.46 &  0.75 &  3.00 &  0.39 &  0.75 &  3.15 \\
    \textbf{\ModelName\ (DN-DAB-DETR)} &  \textbf{0.65} & \textbf{0.87} & \textbf{1.51}  &  \textbf{0.68} &  \textbf{0.89} &  \textbf{1.03} \\
    \hline
    Based on Perplexity & & & &  & & \\
    \hspace{20pt}\emph{Phi-1.5~\citep{li2023textbooks} Perpl.} + GB regressor & 0.17 & 0.45 & 6.11 & 0.32 & 0.71 & 3.07 \\
     \hspace{20pt}\emph{Phi-1.5~\citep{li2023textbooks} Perpl.} + LR classifier & 0.27 & 0.50 & 11.9 & 0.47 & 0.73 & 4.77 \\
    \hline
     Based on TDA & & & &  & & \\
     \hspace{20pt}PHD + TS ML~\citep{kushnarevaai} & 0.24 & 0.46 & 14.40 & 0.17 & 0.36 & 14.45 \\
     \hspace{20pt}TLE + TS Binary~\citep{kushnarevaai} & 0.13 & 0.30 & 22.23 & 0.20 & 0.35 & 18.52  \\
    \hline
    Human baseline~\citep{cutler2021automatic} & 0.23 & 0.40 & 13.88 & - & - & -\\
    \hline
\end{tabular}
\label{tab:RoFT_results}
\end{table*}

\begin{table*}[h]
\centering
\setlength{\tabcolsep}{0.15em}
\caption{
Accuracy for leave-one-out cross-domain evaluation on \emph{RoFT-ChatGPT}. The results for all methods, except ours, were taken from \citet{kushnarevaai}.
}
\small
\begin{tabular}{lllllll}
\hline
& \multicolumn{2}{c}{} &
  \textbf{Pres. } &
  \textbf{Recipes} &
  \textbf{New York} &
  \textbf{Short} \\ %
\textbf{Pred.} &
  \textbf{Model} &
  \textbf{Context} & \textbf{Speeches} &  & \textbf{Times} & \textbf{Stories} \\ 
  \hline
  Text &
  \textbf{\ModelName\ (DN-DAB-DETR)} &
  global & 0.50 & \textbf{0.33} & \textbf{0.55} & \textbf{0.64} \\
  Text &
    RoBERTa SEP~\citep{cutler2021automatic}&
  global & 0.31  & 0.13  & 0.38  & 0.29  \\
Text &
  RoBERTa~\citep{liu2019roberta} &
  global & 0.36 & 0.15 & 0.38 & 0.36 \\
  
  Perpl. & Phi1.5~\citep{li2023textbooks}, GB &
  sent. & \textbf{0.52} & 0.24 & 0.46 & 0.56 \\
Perpl. & Phi1.5~\citep{li2023textbooks},
  LR &
  sent. & 0.41 & 0.21 & 0.45 & 0.52 \\
  PHD &
  TS multi~\citep{kushnarevaai} &
  100 tkn & 0.13 & 0.20 & 0.17 & 0.18  \\
TLE &
  TS Binary~\citep{kushnarevaai} &
  20 tkn & 0.15 & 0.16 & 0.17 & 0.11 \\
\hline
\end{tabular}
\label{tab:cross_RoFT}
\end{table*}

\begin{table}[h!]
    \centering
    \caption{Evaluation of \ModelName\ (DN-DAB-DETR) on RoFT and RoFT-GPT datasets using mAP@0.5-0.95. The table compares the leave-one-out cross-domain setting against models trained on all domains combined.}
    \small
    \begin{tabular}{lc}
        \hline
        \textbf{Dataset} & \textbf{mAP@0.5-0.95} \\
        \hline
        RoFT-ChatGPT Short Stories & 0.7626 \\
        RoFT-ChatGPT Recipes & 0.6046 \\
        RoFT-ChatGPT Pres Speeches & 0.5933 \\
        RoFT-ChatGPT New York Times & 0.7034 \\
        \hline
        RoFT-ChatGPT All domains & 0.8135 \\
        RoFT All domains & 0.7972 \\
        \hline
    \end{tabular}
    \label{tab:RoFT_map_results}
\end{table}

\begin{table}[h]
    \centering
    \setlength{\tabcolsep}{2pt}
        \caption{Boundary detection results (F1@3) on the TriBERT~\citep{zeng2024towards} dataset. \#Bry denotes the number of ground-truth boundaries in the texts. Measurements are presented in original and rescaled formats.}
    \small
    \begin{tabular}{lcccc}
    \hline
        \bf{Methods} & \bf{\#Bry=1} & \bf{\#Bry=2} & \bf{\#Bry=3} & \bf{All} \\
        \hline
        \multicolumn{5}{c}{Original values}\\
        \hline
        TriBERT (p=2) & \bf{0.455} & 0.692 & 0.622 & 0.575 \\
        
        \bf \makecell[l]{\ModelName{}\\(DN-DAB-DETR)} & 0.444 & \bf{0.693} & \bf{0.801} & \bf{0.646} \\
        \hline
        \multicolumn{5}{c}{Rescaled values} \\
        \hline
        TriBERT (p=2) & \bf{0.910} & 0.865 & 0.622 & - \\
        
        \bf \makecell[l]{\ModelName{}\\(DN-DAB-DETR)}  & 0.888 & \bf{0.867} & \bf{0.801} & - \\
    \hline
    \end{tabular}
    \label{tab:tribert_results}
\end{table}

\textbf{RoFT and RoFT-ChatGPT results.} In experiments on the RoFT and RoFT-ChatGPT datasets, we fine-tuned Mistral-7B to distinguish between human-written texts and texts co-written with LLMs. Features from the model's last layer were used to train the \texttt{\ModelName{}~(DN-DAB-DETR)} model. Since each text in these datasets contains at most one human-to-machine transition, the detector uses a single learnable query ($N{=}1$).

\texttt{\ModelName{}~(DN-DAB-DETR)} natively predicts continuous character-level intervals end-to-end, without any heuristic post-processing. Since the official RoFT metrics operate on sentence boundaries, we apply a deterministic character-to-sentence projection solely for evaluation purposes (details in Appendix~\ref{apdx:mapping}).

Table~\ref{tab:RoFT_results} shows that \texttt{\ModelName{}~(DN-DAB-DETR)} beats the RoBERTa baseline by 15\% on RoFT and 13\% on RoFT-ChatGPT, and reduces MSE on RoFT-ChatGPT by a factor of 3.
Table~\ref{tab:cross_RoFT} shows cross-domain results on RoFT-ChatGPT, where models trained on three domains and tested on the fourth. Our approach achieves the best cross-domain generalization, though performance on the \textit{Recipes} domain remains relatively low.

We additionally report the standard mean Average Precision (mAP) adapted for one-dimensional intervals (Table~\ref{tab:RoFT_map_results}). An interval is considered a true positive if its IoU with a ground-truth interval exceeds a given threshold; mAP@0.5:0.95 averages over thresholds from 0.5 to 0.95. Unlike the sentence-level metrics above, mAP operates directly on character-level predictions and requires no projection, confirming that the model achieves strong localization at the native output granularity.

Examples of raw model output on RoFT-ChatGPT are provided in Appendix~\ref{apdx:examples}.

\textbf{TriBERT results}. TriBERT texts contain up to three authorship
boundaries, yielding denser spans; accordingly,
the detector uses 18 learnable queries ($N{=}18$)
to provide sufficient capacity. Because the TriBERT dataset is small, we keep Mistral-7B-v0.3 frozen and feed its embeddings to \texttt{\ModelName{} (DN-DAB-DETR)}. The detector outputs character spans, which we map to sentence boundaries to compute $F1@3$ (Eq.~\ref{eq:eq3}; mapping details in Appendix~\ref{apdx:mapping}).

Results are reported by boundary count (1, 2, 3) and for the full set. With $K\neq3$ the ideal $F1@3$ scores are 0.5, 0.8, 1.0 \citep{zeng2024towards}. We rescale them to a common scale, where the ideal $F1@3$ is 1.0, for clarity. 
Table \ref{tab:tribert_results} shows a 7.1\% gain over TriBERT model on the full set and higher scores for 2- and 3-boundary texts, while performance is similar for the 1-boundary group. Unlike TriBERT, our model stays stable as the number of boundaries increases.

\subsection{Classification Results}
\label{sec:classification}

We fine-tuned Mistral-7B v0.3 with LoRA on five datasets, comparing to baselines provided by the authors of these datasets. All our models were trained on the same training sets used by the authors.

Tables \ref{tab:tweepfake_results} and \ref{tab:turingbench_results} show strong results on \textbf{TweepFake} and \textbf{TuringBench}, outperforming statistical methods and fine-tuned LM baselines across diverse domains and generators. 

\textbf{MAGE results.} Table \ref{tab:mage_results} compares \texttt{\ModelName{} (Mistral-7B)} with the strongest baseline reported by the dataset authors (full results in Appendix~\ref{apdx:mage_comp}) and shows that our model reaches AUROC = 0.99 and AvgRec = 0.96 on the full large-scale split. It keeps strong generalisation: AvgRec = 0.89 in the \emph{unseen-domain + unseen-model} test, 0.69 under paraphrase attacks, and AUROC = 0.98 / AvgRec = 0.92 in the \textit{out-of-model} setting, where texts from specific generators were excluded during training.

\textbf{Effect of backbone size.} To gauge the impact of scale we repeated the full-data experiment on MAGE (the largest corpus in our experiments) using three larger LoRA-tuned backbones: Mistral-Nemo-Base-2407 (12 B), Mistral-Small-24B-Base-2501 (24 B), and Qwen2.5-72B-Instruct (72 B). 
As reported in Table \ref{tab:mage_backbones}, accuracy rises with backbone size overall, yet the 72B Qwen variant drops to the lowest score, hinting at overfitting. Because the gains beyond 7B are modest relative to the added compute, we keep the 7 B backbone for all other datasets; it trains quickly, fits standard memory limits, and is less prone to overfitting on small corpora even with LoRA.

\textbf{In summary}, our approach with 7B backbone effectively distinguishes LLM-generated texts from human-written ones when trained on both small and large datasets. The experiments demonstrate the robustness of our method for out-of-domain and out-of-model detection, as well as its resistance to paraphrasing attacks. Additionally, Appendix~\ref{apdx:zero_shot_comp} 
presents a comparison between the fine-tuned \texttt{\ModelName{}~(Mistral-7B)} models and the Mistral-7B-Instruct-v0.3 model, evaluated in a zero-shot setting across each test set.

\begin{table}[!ht]
		\centering
        \caption{Experimental results on TweepFake test set. F1 scores are reported as 'human' / 'machine'. }
        \small
		\begin{tabular}{lccc}
		\hline
			\bf {Method} & \bf{F1} & \bf{Acc}\\ 
            \hline
			BERT~\citep{devlin2018bert} & 0.890 / 0.892 & 0.891\\
			DistilBERT~\citep{sanh2019distilbert} & 0.886 / 0.888 & 0.887\\
			RoBERTa~\citep{liu2019roberta} & 0.895 / 0.897 & 0.896\\
			XLNet~\citep{yang2019xlnet} & 0.871 / 0.882 &  0.877\\
                \bf {\ModelName\ (Mistral-7B)} & \textbf{0.944 / 0.942} & \textbf{0.943} \\
		\hline
		\end{tabular}
            \label{tab:tweepfake_results}
\end{table}

\begin{table}[!ht]
		\centering
        \caption{
			 Experimental results (F1) on two TuringBench subsets. F1 is calculated for the machine-generated category.}
        \small
		\begin{tabular} {p{3.2cm} >{\centering\arraybackslash}p{1.5cm} >{\centering\arraybackslash}p{1.5cm}}
			\hline
			\bf {Method} & \bf{FAIR\_wmt20} &  \bf{GPT-3}\\ 
            \hline
			GLTR \citep{gltr} & 0.4907 & 0.3476\\
			BERT~\citep{devlin2018bert} & 0.4701 & 0.7944\\
			RoBERTa~\citep{liu2019roberta} & 0.4531 & 0.5209\\
                \textbf{\ModelName\ (Mistral-7B)} & \textbf{0.9966} & \textbf{0.9709} \\
			\hline
		\end{tabular}
            \label{tab:turingbench_results}
\end{table}

\begin{table}
\centering
\caption{Classification performance on MAGE dataset in different scenarios including performance on the two challenging test sets. 
To test on challenging test sets (Unseen Domains \& Unseen Model, Paraphrasing Attack) the model trained on Arbitrary-domains \& Arbitrary-models dataset was used. Metrics for the Longformer~\citep{beltagy2020longformer} method was taken from the authors of MAGE dataset.}
\small
\begin{tabular}{lcccc}
\hline
\textbf{Methods} & \textbf{AvgRec} & \textbf{AUROC} \\
\hline
\multicolumn{3}{c}{Arbitrary-domains \& Arbitrary-models }\\
\hline
Longformer & 0.91 & 0.99 \\
\textbf{\ModelName\ (Mistral-7B)} & \textbf{0.96} & 0.99 \\
\hline
\multicolumn{3}{c}{Unseen Domains \& Unseen Model}\\
\hline
Longformer & 0.76& 0.94 \\
\textbf{\ModelName\ (Mistral-7B)} & \textbf{0.89} & 0.96 \\
\hline
\multicolumn{3}{c}{Paraphrasing Attack}\\
\hline
Longformer & 0.67 & 0.75 \\
\textbf{\ModelName\ (Mistral-7B)} & \textbf{0.69} & 0.74 \\
\hline
\multicolumn{3}{c}{Out-of-distribution Detection: Unseen models} \\
\hline
Longformer & 0.87 & 0.95 \\
\textbf{\ModelName\ (Mistral-7B)} & \textbf{0.92} & 0.98 \\
\hline
\end{tabular}
\label{tab:mage_results}
\end{table}

\begin{table}
\centering
\caption{Impact of backbone size on MAGE full set.}
\small
\begin{tabular}{lcccc}
\hline
\textbf{Model} & \textbf{AvgRec} & \textbf{AUROC} \\
\hline
\ModelName\ (Mistral-7B) & 0.9611 & 0.9923 \\
\ModelName\ (Mistral-12B) & 0.9630 & \textbf{0.9941} \\
\ModelName\ (Mistral-24B) & \textbf{0.9685} & 0.9937 \\
\ModelName\ (Qwen-72B) & 0.8338 & 0.9697 \\
\hline
\end{tabular}
\label{tab:mage_backbones}
\end{table}

%% file: latex/6_Conclusions.tex
\section{Conclusions}

We presented \texttt{\ModelName}, a unified framework that combines a LoRA-tuned backbone LLM with two lightweight heads:  
(i) a DN-DAB-DETR module for precise character-level localization of LLM-generated spans, and  (ii) a streamlined MLP for document-level authorship verification.

Our experiments on three Human–Machine collaborative datasets demonstrate that DETR-style transformers can be successfully translated from computer vision to the textual domain, treating generated spans as discrete objects to achieve high-fidelity localization. Simultaneously, the shared backbone matches or surpasses prior baselines on three binary-classification corpora, confirming that the learned representations are both \emph{robust} and \emph{transferable} across tasks of varying granularity.

Crucially, unlike methods constrained by sentence boundaries or explicit document structures, \texttt{\ModelName} offers flexible, boundary-free detection. It operates effectively without predefined segmentation, showing strong generalization capabilities across diverse setups (from pre-trained to fine-tuned backbones) and in challenging out-of-domain scenarios.

\section{Limitations}
\label{sec:limitations}

\noindent \textbf{Context Window Constraints.} To optimize computational efficiency during training, we explicitly restrict the input sequence length, although the backbone supports longer contexts. Consequently, documents exceeding this limit are processed in independent chunks, potentially obscuring long-range dependencies across segment boundaries. However, this is a hyperparameter choice; the core architecture scales naturally to larger context windows given sufficient computational resources.

\noindent \textbf{Language Scope.} This study is intentionally scoped to English to ensure rigorous comparison with established benchmarks. Since the unified backbone is multilingual by design, extending \ModelName\ to other languages requires no architectural modifications, only the curation of appropriate training data.

\noindent \textbf{Backbone Dependency.} We report results using Mistral-7B due to its favourable quality-to-compute trade-off. However, the pipeline is model-agnostic; the framework permits swapping the backbone for any decoder-style LLM (e.g., LLaMA, Qwen) to adapt to specific resource constraints or domain requirements.

\noindent \textbf{Benchmark Saturation.} Near-perfect scores on smaller corpora like TuringBench may reflect their limited diversity rather than unsolved challenges. In datasets with few source domains and generator models, distinct artifacts persist, simplifying detection~\citep{gritsai2024ai}. Thus, these results may overstate real-world performance. To address this limitation, in concurrent work we assembled a substantially larger and more diverse benchmark and evaluated GigaCheck on it~\citep{llmtrace}.

\section{Ethical Statement}
\paragraph{Interpretability and Misuse.} While \ModelName\ improves transparency by localising specific AI-generated spans rather than providing a black-box document-level verdict, it does not achieve perfect accuracy. Performance can fluctuate based on the generator model, text length, and domain. Consequently, the detector should be used as an \textit{assistive tool} for human verification, not as the sole basis for high-stakes decisions (e.g., academic disciplinary actions). We disclaim responsibility for any reputational damage or adverse consequences arising from the unverified reliance on its outputs.

%% file: latex/7_Appendix.tex
\section{Pre-trained VS fine-tuned models' embeddings}
\label{apdx:pretrained_ft}

Table \ref{tab:RoFT_results_additional} presents a comparison of detection model performance on the RoFT and RoFT-ChatGPT datasets using two different setups. In the first experiment, we fine-tuned the Mistral-7B model to perform a text classification task with two labels: \textit{'Human'} and \textit{'AI-Human Collaborative'}, and used this model to extract text features for DETR model training. In the second experiment, we utilized the pre-trained Mistral-7B v0.3 model for feature extraction. Two DN-DAB-DETR models were then trained using these two types of features. The results indicate that the detection model performs better with features from the fine-tuned model; however, the model trained with text representations from the pre-trained model also achieves strong results on both datasets. We also provide results from \citet{kushnarevaai} for comparison.

\begin{table*}[!t]
  \centering\setlength{\tabcolsep}{4pt}
    \caption{Boundary detection results on RoFT and RoFT-ChatGPT datasets. `\dag' denotes the DETR model was trained on text features from pre-trained Mistral-7B v0.3 model. \textbf{Bold} shows the best method, \underline{underlined} - second best.}
  \small
  \begin{tabular}{lllllll}
  \hline
    \textbf{Method} & \multicolumn{3}{c}{\textbf{RoFT}} & \multicolumn{3}{c}{\textbf{RoFT-ChatGPT}} \\
    & Acc & SoftAcc1 & MSE & Acc & SoftAcc1 & MSE \\
    \hline
    RoBERTa + SEP & 49.64 \% & 79.71 \% & \underline{2.63} & \underline{54.61} \% & 79.03 \% & 3.06 \\
    RoBERTa &  46.47 \% &  74.86 \% &  3.00 &  39.01 \% &  75.18 \% &  3.15 \\
    \textbf{\ModelName\ (DN-DAB-DETR)}\dag & \underline{60.10} \% &  \underline{81.48} \% &  2.77 &  51.37 \% &  \underline{80.12} \% & \underline{1.93} \\
    \textbf{\ModelName\ (DN-DAB-DETR)} &  \textbf{64.63} \% & \textbf{86.68}  \% & \textbf{1.51}  &  \textbf{67.65} \% &  \textbf{88.98} \% &  \textbf{1.03} \\
    \hline
\end{tabular}
\label{tab:RoFT_results_additional}
\end{table*}

\section{Hyperparameters and experimental setup}
\label{apdx:hyperparameters}

We fine-tune Mistral-7B-v0.3\footnote{https://huggingface.co/mistralai/Mistral-7B-v0.3} for a binary classification task to distinguish between human-written and machine-generated content using LoRA. Models training were done using Hugging Face Transformers\footnote{https://github.com/huggingface/transformers} with bfloat16 precision. LoRA settings via the PEFT\footnote{https://github.com/huggingface/peft} library include: $r = 8$, \textit{lora\_alpha} = 16, \textit{lora\_dropout} = 0.1, and $bias = "none"$. Only query and value projection matrices in attention modules were adapted. We used AdamW~\citep{decoupled} with a cosine learning rate scheduler~\citep{sgdr}. The DETR model’s encoder and decoder each had 3 layers. The loss weights were set to 10.0 for L1, 1.0 for gIoU, 4.0 for Focal Loss, 9.0 for denoised L1, and 3.0 for denoised gIoU. 

During training, we augmented the data by randomly selecting between \textit{'minimum sequence length'} to \textit{'maximum sequence length'} tokens from each text. To optimize the models, we used the AdamW optimizer with a cosine learning rate schedule and also applied a weight for the \textit{'Human'} category in the cross-entropy function. The dataset-specific hyperparameters used for the experiments are listed in the table \ref{tab:classification_hyperparams}. 

When training a detection model to find LLM-generated intervals in text, we follow three steps: 1) fine-tune the Mistral-7B model on two or three categories, 2) extract features for the dataset from the trained model, 3) train the DETR model using extracted features as input data. The training is divided into three stages, firstly because this significantly speeds up the training process, and secondly because LLM and DETR models converge at different rates.

To train DN-DAB-DETR models, we also used the AdamW optimizer with a cosine learning rate schedule. During training we did not apply any text augmentations. The number of learnable queries $N$ reflects the maximum span density per text in each dataset (see Section~\ref{sec:detection}). The dataset-specific hyperparameters used for the experiments are listed in the table \ref{tab:detection_hyperparams}.

\begin{table*}[!t]
    \centering
    \caption{
    Hyperparameters for the classification experiments.
    }
    \small
    \begin{tabular}{lccccc}
        \hline
        \bf {Parameter} & \bf{MAGE} &  \bf{TuringBench} & \bf{TweepFake} \\ 
        \hline
        \makecell{max sequence length} & 1024 & 1024 & 1024\\
        \makecell{minimum sequence \\length for augmentatoins} & 900 & 15 & 900\\
            \makecell{train batch size} & 64 & 32 & 32\\
            \makecell{gradient accumulation \\steps} & 1 & 2 & 2\\
            \makecell{learning rate} &  3e-4 & 3e-4 & 3e-4\\
            \makecell{cross entropy weight for \\human category} & 2 & 1 & 1\\
            \makecell{num train epochs} & 3 & 5 & 4\\
            \makecell{GPUs} & \makecell{1xNvidia \\H100} & \makecell{1xNvidia \\H100} & \makecell{1xNvidia \\H100}\\
            \makecell{the fine-tuning time} & 48h & 2h & 2h \\
    \hline
    \end{tabular}
    \label{tab:classification_hyperparams}
\end{table*}

\begin{table*}[!t]
		\centering
      \caption{
    Hyperparameters for the span-detection (DN-DAB-DETR) experiments.}
        \small
		\begin{tabular}{lccc}
		\hline
			\bf {Parameter} & \bf{RoFT} &  \bf{RoFT-ChatGPT} & \bf{TriBERT}\\ 
            \hline
                \makecell{number of queries} & 1 & 1 & 18 \\
			\makecell{max sequence length} & 512 & 512 & 1024 \\
                \makecell{train batch size} & 32 & 32 & 64\\
                \makecell{gradient accumulation \\steps} & 2 & 2 & 1 \\
                \makecell{learning rate} & 1e-4 & 1e-4 & 2e-4 \\
                \makecell{num train epochs} & 75 & 75 & 75 \\
                \makecell{GPUs} & \makecell{1xNvidia \\H100} & \makecell{1xNvidia \\H100} & \makecell{1xNvidia \\H100}\\
                \makecell{the DETR training time} & 5h & 3h & 6h  \\
                \makecell{the Mistral fine-tuning time} & 3h & 2h & (without fine-tuning) \\
		\hline
		\end{tabular}
  \label{tab:detection_hyperparams}
\end{table*}

\section{MAGE comparison}
\label{apdx:mage_comp}

Table \ref{tab:mage_results_all} shows the results of comparing \ModelName\ with Mistral-7B with all detectors considered by the authors of the MAGE dataset. We also report \ModelName’s performance on the MAGE full set (Arbitrary-domains \& Arbitrary-models) using backbones of different sizes. We fine-tuned three large backbones: Mistral-Nemo-Base-2407\footnote{https://huggingface.co/mistralai/Mistral-Nemo-Base-2407} (12B), Mistral-Small-24B-Base-2501\footnote{https://huggingface.co/mistralai/Mistral-Small-24B-Base-2501} (24B), and Qwen2.5-72B-Instruct\footnote{https://huggingface.co/Qwen/Qwen2.5-72B-Instruct} (72B).

\begin{table*}[t!]
\centering
\caption{Classification performance on MAGE dataset in different scenarios including performance on the two challenging test sets. 
To test on challenging test sets the model trained on Arbitrary-domains \& Arbitrary-models dataset was used.}
\small
\begin{tabular}{lcccc}
\hline
\textbf{Methods} & \textbf{HumanRec} & \textbf{MachineRec} & \textbf{AvgRec} & \textbf{AUROC} \\
\hline
\multicolumn{5}{c}{Arbitrary-domains \& Arbitrary-models }\\
\hline
FastText~\citep{joulin2016bag} & 86.34\% & 71.26\% & 78.80\% & 0.83 \\
GLTR~\citep{gltr} & 12.42\% & 98.42\% & 55.42\% & 0.74 \\
DetectGPT~\citep{detectgpt} & 86.92\% & 34.05\% & 60.48\% & 0.57 \\
Longformer~\citep{beltagy2020longformer} & 82.80\% & 98.27\% & 90.53\% & 0.99 \\
\textbf{\ModelName\ (Mistral-7B)} & 95.72\% & 96.49\% & 96.11\% & 0.99 \\
\textbf{\ModelName\ (Mistral-12B)} & 95.29\% & 97.32\% & 96.30\% & 0.99 \\
\textbf{\ModelName\ (Mistral-24B)} & 96.94\% & 96.76\% & \textbf{96.85}\% & 0.99 \\
\textbf{\ModelName\ (Qwen-72B)} & 83.38\% & 96.62\% & 83.38\% & 0.97 \\
\hline
\multicolumn{5}{c}{Unseen Domains \& Unseen Model}\\
\hline
FastText~\citep{joulin2016bag} & 71.78\% & 68.88\% & 70.33\% & 0.74 \\
GLTR~\citep{gltr} & 16.79\% & 98.63\% & 57.71\% & 0.73 \\
Longformer~\citep{beltagy2020longformer} & 52.50\% & 99.14\% & 75.82\% & 0.94 \\
\textbf{\ModelName\ (Mistral-7B)} & 79.71\% & 97.38\% & \textbf{88.54\%} & 0.96 \\
\hline
\multicolumn{5}{c}{Paraphrasing Attack}\\
\hline
FastText~\citep{joulin2016bag} & 71.78\% & 50.00\% & 60.89\% & 0.66 \\
GLTR~\citep{gltr} & 16.79\% & 82.44\% & 49.61\% & 0.47 \\
Longformer~\citep{beltagy2020longformer} & 52.16\% & 81.73\% & 66.94\% & 0.75 \\
\textbf{\ModelName\ (Mistral-7B)} & 79.66\% & 58.24\% & \textbf{68.95\%} & 0.74 \\
\hline
\multicolumn{5}{c}{Out-of-distribution Detection: Unseen models} \\
\hline
FastText~\citep{joulin2016bag}  & 83.12\% & 54.09\% & 68.61\% & 0.74 \\ 
GLTR~\citep{gltr}  & 25.77\% & 89.21\% & 57.49\% & 0.65 \\  
DetectGPT~\citep{detectgpt} & 48.67\% & 75.95\% & 62.31\% & 0.60 \\ 
Longformer~\citep{beltagy2020longformer} & 83.31\% & 89.90\% & 86.61\% & 0.95 \\
\textbf{\ModelName\ (Mistral-7B)} & 95.65\% & 89.00\% & \textbf{92.32\%} & 0.98 \\
\hline
\end{tabular}
\label{tab:mage_results_all}
\end{table*}

\section{Mistral-7B-v0.3 zero-shot classification results}
\label{apdx:zero_shot_comp}

Table \ref{tab:zero_shot_results} presents the results of comparing \ModelName\ with Mistral-7B fine-tuned with LoRA on five classification datasets against the Mistral-7B-Instruct-v0.3\footnote{https://huggingface.co/mistralai/Mistral-7B-Instruct-v0.3} model, evaluated in a zero-shot setting. The comparison was conducted on the test sets.

\begin{table*}[h]
		\centering
        \caption{
			 Experimental results (F1 scores) on the test sets for classification datasets. F1 is calculated for the machine-generated category. We compare the Mistral-7B-Instruct-v0.3 model evaluated in a zero-shot setting with fine-tuned Mistral-7B-v0.3 models.}
	\small	
    \begin{tabular}{lcccc}
			\hline
			\bf {Method} & \bf{TweepFake} &  \bf \makecell{TuringBench \\ FAIR\_wmt20} &  \bf \makecell{TuringBench \\ GPT-3} & \bf{MAGE} \\
            \hline
			Mistral-7B-Instruct-v0.3 & 0.640 & 0.537 & 0.500 & 0.633  \\
            \textbf{\ModelName\ (Mistral-7B)} & 0.942 & 0.997 & 0.971 & 0.96 \\
			\hline
		\end{tabular}
            \label{tab:zero_shot_results}
\end{table*}

\section{Evaluation metrics for detection datasets}
\label{apdx:metrics}

For each detection dataset, we compute specific metrics.

Followed the approach of the authors in \citet{kushnarevaai}, we compute mean squared error \textbf{MSE}$=\frac{1}{N}\sum_{i=1}^{N}{(y_i-\hat{y_i})^2}$ between the predicted boundaries $\hat{y}$ and the true boundaries $y$, where a boundary is the sentence number at which authorship in the text changes from human to LLM, and $N$ represents the number of samples. It is worth noting that in both datasets from \citet{kushnarevaai}, each text contains no more than one boundary. The authors also propose reporting accuracy (\textbf{Acc}) of boundary detection and soft accuracy (\textbf{SoftAcc1}), the proportion of predictions that are off from the correct label by no more than one.

Finally, the authors of \citep{wang2024m4gt} evaluate model prediction quality using the mean absolute error \textbf{MAE}$=\frac{1}{N}\sum_{i=1}^{N}{|y_i-\hat{y_i}|}$, where $\hat{y}$ denotes the predicted word number that separates human and AI-generated parts of the text, $y$ represents ground-truth word number, and $N$ is the number of samples. The problem statement in \citep{wang2024m4gt} implies that there is only one such word boundary per text.

F1@K metric proposed by \citet{zeng2024towards} to asses the performance of model in boundaries detection task is described in Eq. \ref{eq:F1@K}. K was set to 3 for all measurements on TriBERT dataset.

\begin{equation}
\label{eq:F1@K}
        F1@K=2\cdot\frac{|L_{topK}\cap L_{Gt}|}{|L_{topK}|+|L_{Gt}|}
\end{equation}

\section{Interval post-processing}
\label{apdx:mapping}

The DETR predictions are post-processed as follows for experiments on the \textbf{RoFT} and \textbf{RoFT-ChatGPT} datasets: let $t_I$ be the start of the interval $I$, and $start_i$, $end_i$ be the indexes of the first and last characters of the \(i\)-th sentence. If the \(i\)-th sentence contains $t_I$, the sentence number $i'$, to which we map DN-DAB-DETR's prediction, is calculated as follows:

\begin{equation}
i' = i +
\begin{cases}
1, & \text{if } t_I \geq \frac{start_i + end_i}{2}, \\
0, & \text{if } t_I < \frac{start_i + end_i}{2}.
\end{cases}
\end{equation}

For the \textbf{TriBERT} experiments, DETR predictions undergo the following post-processing steps: let $b_i$ and $b_{i+1}$ denote the beginnings of the $n$ and $n+1$ sentences in characters and let $p_j$ denote the beginning or the end of the predicted interval in characters. Then the boundary $B$ for $p_j$ is calculated as:

\begin{equation}
\label{eq:eq_4}
B(p_j) =
\begin{cases}
b_i & \text{if } p_j < \frac{b_i + b_{i+1}}{2}, \\
b_{i+1} & \text{if } p_j \geq \frac{b_i + b_{i+1}}{2}. \\
\end{cases}
\end{equation}

Therefore, if the predicted start or end of the interval falls in the first half of sentence $n$, we map it to the beginning of sentence $n$. If it falls in the second half, we map it to the beginning of the next sentence, $n+1$. As a result, each boundary determines the sentence number where the text's authorship changes. Note that if a boundary is equal to the beginning or the end of the whole text, we remove it, since a boundary can only be between two sentences.

\section{Examples of the DETR model output}
\label{apdx:examples}

Tables \ref{tab:examples1} and \ref{tab:examples2} present examples of work of the model trained on the RoFT-ChatGPT dataset. Table \ref{tab:examples1} shows the ground truth and output result for test samples from the \textit{'Short Stories'} and \textit{'New York Times'} domains. Table \ref{tab:examples2} shows the ground truth and output result for test samples from the \textit{'Recipes' } and \textit{'Presidential Speeches'} domains.

\begin{table*}[!ht]
    \centering
    \caption{Examples from the test set of the raw model's output, trained on the RoFT-ChatGPT dataset. \textbf{Bold} text indicates either the ground truth interval or the predicted one.}
    \resizebox{\textwidth}{!}{%
    \begin{tabular}{lp{12cm}}
        \hline
        Domain: \textit{Short Stories}\\
        \hline
        \begin{minipage}{\textwidth}
        \textbf{GT:} Aryton blinked and rubbed his head. It had been a very high speed crash. He expected the impact to hurt more, but the whole thing just felt quite... fuzzy. There didn't seem to be any track marshals around, which was odd, Aryton looked back towards the corner where he'd lost control. Nothing there, he pulled himself out of the car and scurried over the crash barrier to safety. That's funny, he thought as he looked back at the crash, the car doesn't seem damaged. \textbf{Aryton walked back towards his car and inspected it closely. It was as if the crash had never happened, there wasn't a scratch on it. He checked the fuel gauge, it was full, and the tires were still warm to the touch. It was a brand new car and one of the fastest ones that he had ever driven.}\\[4pt]
        \textbf{Output:} Aryton blinked and rubbed his head. It had been a very high speed crash. He expected the impact to hurt more, but the whole thing just felt quite... fuzzy. There didn't seem to be any track marshals around, which was odd, Aryton looked back towards the corner where he'd lost control. Nothing there, he pulled himself out of the car and scurried over the crash barrier to safety. That's funny, he thought as he looked back at the crash, the car doesn't seem da\textbf{maged. Aryton walked back towards his car and inspected it closely. It was as if the crash had never happened, there wasn't a scratch on it. He checked the fuel gauge, it was full, and the tires were still warm to the touch. It was a brand new car and one of the fastest ones that he had ever drive}n.\\
        \end{minipage} \\
        \hline
        Domain: \textit{New York Times} \\
        \hline 
        \begin{minipage}{\textwidth}
        \textbf{GT:} 
        ... For many in the industry, it was the final seal of approval on a technology that remained controversial as long as it was exclusive to smaller, less conservative computer makers. But that interpretation does not sit well with Irving Wladawsky-Berger, who is responsible for the supercomputing business at the International Business Machines Corporation. " For me to say now we've finally put our seal of approval on this would sound supremely arrogant," he said. " Let's just say we have committed to build a product family of parallel RISC systems that scale up from our RS/6000." RISC, or reduced instruction set computing, is a technology that speeds processing by relegating more tasks to software; the RS/6000 is the name for both a chip set and a computer work station produced by I.B.M. using RISC. Dr. Wladawsky-Berger said the impetus to create a massively parallel supercomputer came from RS/6000 customers who were creating a sort of virtual parallel processor by linking multiple work stations. " There were people pushing at I.B.M., but they were pushing in many different directions," he said. " Supercomputing is an area where if you get seven smart people together, you get 17 different architectures." " \textbf{But," he added, "we knew we had to do something because we were seeing more and more of our customers doing this and we knew we had to provide them with a scalable solution.} \\[4pt]
        \textbf{Output:} ... For many in the industry, it was the final seal of approval on a technology that remained controversial as long as it was exclusive to smaller, less conservative computer makers. But that interpretation does not sit well with Irving Wladawsky-Berger, who is responsible for the supercomputing business at the International Business Machines Corporation. " For me to say now we've finally put our seal of approval on this would sound supremely arrogant," he said. " Let's just say we have committed to build a product family of parallel RISC systems that scale up from our RS/6000." RISC, or reduced instruction set computing, is a technology that speeds processing by relegating more tasks to software; the RS/6000 is the name for both a chip set and a computer work station produced by I.B.M. using RISC. Dr. Wladawsky-Berger said the impetus to create a massively parallel supercomputer came from RS/6000 customers who were creating a sort of virtual parallel processor by linking multiple work stations. " There were people pushing at I.B.M., but they were pushing in many different directions," he said. " Supercomputing is an area where if you get seven smart people together, you get 17 different architectures." " But," he added, "we \textbf{knew we had to do something because we were seeing more and more of our customers doing this and we knew we had to provide them with a scalable soluti}on. \\
        \end{minipage} \\
        \hline
    \end{tabular}%
}
    \label{tab:examples1}
\end{table*}

\begin{table*}[!ht]
    \centering
    \caption{Examples from the test set of the raw model's output, trained on the RoFT-ChatGPT dataset. \textbf{Bold} text indicates either the ground truth interval or the predicted one.}
    \resizebox{\textwidth}{!}{%
    \begin{tabular}{lp{12cm}}
        \hline
        Domain: \textit{Recipes} \\
        \hline 
        \begin{minipage}{\textwidth}
        \textbf{GT:} HOW TO MAKE: Make-Ahead Turkey Gravy Ingredients: 2 tablespoons canola oil 2 lbs turkey wings 1 cup dry white wine 3 tablespoons olive oil 1 medium yellow onion, halved 2 carrots, cut in 2 inch pieces 2 celery ribs, cut in 2 inch pieces plus a handful of the celery leaves 1 head garlic, cut in half 2 sprigs fresh thyme 2 sprigs fresh sage 2 sprigs fresh rosemary 10 black peppercorns 2 bay leaves 6 cups low sodium chicken broth 8 tablespoons flour 4 tablespoons butter, if needed 12 teaspoon white vinegar Kitchen Bouquet, if desired. \textbf{Instructions: 1. Preheat the oven to 375F.2. In a large roasting pan, toss the turkey wings with canola oil.3. Roast the turkey wings for about 1 hour, or until deeply golden brown.4. Transfer the turkey wings to a large pot and pour in the white wine.5. Over medium-high heat, bring to a simmer and scrape up any browned bits from the bottom of the roasting pan.6. Simmer for about 5 minutes, or until the wine has reduced by half.7. Pour the wine mixture over the turkey wings and set aside.8. In a large skillet, heat the olive oil over medium heat.9.}
        \\[4pt]
        \textbf{Output:} HOW TO MAKE: Make-Ahead Turkey Gravy Ingredients: 2 tablespoons canola oil 2 lbs turkey wings 1 cup dry white wine 3 tablespoons olive oil 1 medium yellow onion, halved 2 carrots, cut in 2 inch pieces 2 celery ribs, cut in 2 inch pieces plus a handful of the celery leaves 1 head garlic, cut in half 2 sprigs fresh thyme 2 sprigs fresh sage 2 sprigs fresh rosemary 10 black peppercorns 2 bay leaves 6 cups low sodium chicken broth 8 tablespoons flour 4 tablespoons butter, if needed 12 teaspoon white vinegar Kitchen Bouquet, if desired\textbf{. Instructions: 1. Preheat the oven to 375F.2. In a large roasting pan, toss the turkey wings with canola oil.3. Roast the turkey wings for about 1 hour, or until deeply golden brown.4. Transfer the turkey wings to a large pot and pour in the white wine.5. Over medium-high heat, bring to a simmer and scrape up any browned bits from the bottom of the roasting pan.6. Simmer for about 5 minutes, or until the wine has reduced by half.7. Pour the wine mixture over the turkey wings and set aside.8. In a large skillet, heat the olive oil over medium heat}.9.\\
        \end{minipage} \\
        \hline
        Domain: \textit{Presidential Speeches} \\
        \hline 
        \begin{minipage}{\textwidth}
        \textbf{GT:} "An Association of Nations" by President Warren G. Harding on July 22, 1920. My countrymen, we believe the unspeakable sorrows, the immeasurable sacrifices, the awakened convictions, and the aspiring conscience of humankind must commit the nations of the earth to a new and better relationship. It need not be discussed now what motives plunged the world into war. It need not be inquired whether we asked the sons of this republic to defend our national rights, as I believe we did, or to purge the Old World of the accumulated ills of rivalry and greed. The sacrifices will be in vain if we cannot acclaim a new order with added security to civilization and peace maintained. One may readily sense the conscience of our America. I am sure I understand the purpose of the dominant group of the Senate. We were not seeking to defeat a world aspiration. \textbf{We were not seeking to withhold our country from doing its part in the world's great work. We were seeking only to safeguard our own sovereignty and to enter into any relationship with other nations only after full and free discussion and deliberation.}\\[4pt]
        \textbf{Output:} "An Association of Nations" by President Warren G. Harding on July 22, 1920. My countrymen, we believe the unspeakable sorrows, the immeasurable sacrifices, the awakened convictions, and the aspiring conscience of humankind must commit the nations of the earth to a new and better relationship. It need not be discussed now what motives plunged the world into war. It need not be inquired whether we asked the sons of this republic to defend our national rights, as I believe we did, or to purge the Old World of the accumulated ills of rivalry and greed. The sacrifices will be in vain if we cannot acclaim a new order with added security to civilization and peace maintained. One may readily sense the conscience of our America. I am sure I understand the purpose of the dominant group of the Senate. We were not seeking to defeat a world aspiration. We were not seeking to withhold our country \textbf{from doing its part in the world's great work. We were seeking only to safeguard our own sovereignty and to enter into any relationship with other nations only after full and free discussion and deliberati}on.\\
        \end{minipage} \\
        \hline
    \end{tabular}%
}
    \label{tab:examples2}
\end{table*}